\documentclass{article}
\usepackage[]{graphicx}\usepackage[]{color}
\makeatletter
\def\maxwidth{ %
  \ifdim\Gin@nat@width>\linewidth
    \linewidth
  \else
    \Gin@nat@width
  \fi
}
\makeatother

\definecolor{fgcolor}{rgb}{0.345, 0.345, 0.345}

\usepackage{framed}
\makeatletter
 {\par\unskip\endMakeFramed%
 \at@end@of@kframe}
\makeatother

\definecolor{shadecolor}{rgb}{.97, .97, .97}
\definecolor{messagecolor}{rgb}{0, 0, 0}
\definecolor{warningcolor}{rgb}{1, 0, 1}
\definecolor{errorcolor}{rgb}{1, 0, 0}
\newenvironment{knitrout}{}{} 

\usepackage{alltt}
\newcommand{\SweaveOpts}[1]{}  
\newcommand{\SweaveInput}[1]{} 
\newcommand{\Sexpr}[1]{}       

\usepackage{spconf,amsmath,graphicx}
\usepackage{latexsym}
\usepackage[pdftex,dvipsnames]{xcolor}  
\usepackage{array}
\usepackage{multirow}
\usepackage{caption}
\usepackage{subcaption}
\usepackage{dcolumn}
\usepackage{cases}
\usepackage{tikz}

\newcolumntype{H}{>{\setbox0=\hbox\bgroup}c<{\egroup}@{}}

\usepackage{expex}
\lingset{glwordalign=center}

\usepackage{xargs}                      

\title{Phonetically-Oriented Word Error Alignment for \\ Speech Recognition Error Analysis in Speech Translation}
\name{Nicholas Ruiz, Marcello Federico}
\address{Fondazione Bruno Kessler \\
Trento, Italy}
\begin{document}

\section{ASR Error Analysis}
\label{sec:asr-eval}
In this set of experiments, we observe the contribution of particular error types to the global WER and POWER scores for each ASR system.
We outline the shortcomings of WER's statistics due to the erratic behavior of Levenshtein aligners.


\begin{knitrout}
\definecolor{shadecolor}{rgb}{0.969, 0.969, 0.969}\color{fgcolor}\begin{figure}

{\centering \includegraphics[width=\maxwidth]{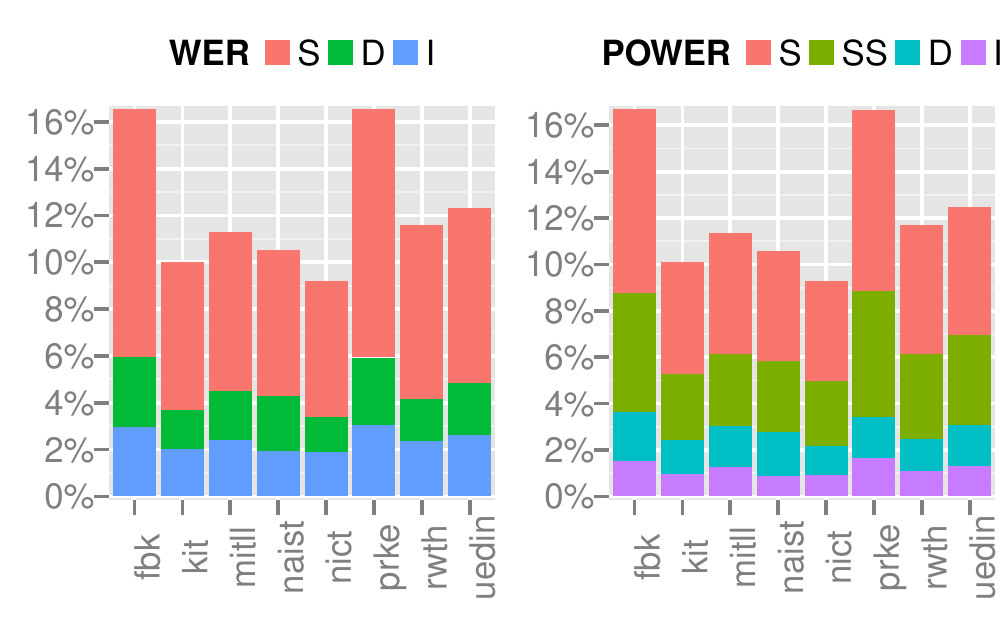} 

}

\caption[Distribution of error types for WER (left) and POWER (right) for each IWSLT 2013 ASR evaluation participant]{Distribution of error types for WER (left) and POWER (right) for each IWSLT 2013 ASR evaluation participant.}\label{fig:asrErrorSummaryPlot}
\end{figure}

\end{knitrout}

\subsection{Basic Levenshtein error types}
\label{sec:asr-eval:basic}

We begin by looking at the basic ASR error types (S, D, I, and SS), which implicitly contain no linguistic information.
Fig.~\ref{fig:asrErrorSummaryPlot} shows the contribution of the basic Levenshtein error types toward the error rate score for each ASR system.
%
%
According to WER, substitutions intuitively make up the majority of error types 
($62.3\% \pm 0.7\%$). 
Across all ASR systems, WER suggests that the number of deletions are slightly lower than the number of insertion errors
($17.9\% \pm 0.7\%$ deletions
and $19.8\% \pm 0.5\%$ insertions).

However, POWER suggests that
roughly half of 
these alleged insertion errors ($10.0\% \pm 0.3\%$)
are instances where a larger reference word is being hypothesized as a homophonic sequence of shorter words.
Likewise, a portion of ``deletion'' errors are instances where multiple reference words were hypothesized as a longer homophonic word
($4.1\% \pm 0.5\%$).
%
%
Since these substitution span errors are typically cases of one-to-many alignments, the number of reported word-level substitution errors are reduced.
As such, POWER claims that
$30.0\% (\pm 0.7\%)$ of the errors are substitution spans involving homophony, 
leaving $13.8\% (\pm 0.8\%)$ of the remaining errors as deletions
and only $9.8\% (\pm 0.3\%)$ as insertions whose pronunciations do not align to any words -- both measures are substantially lower than those reported by WER.
The remaining
$46.4\% (\pm 0.5\%)$ 
are word-level substitutions.

\begin{table}[tb]
\centering
{\footnotesize
\begin{tabular}{l|rr|rrr}
  \hline
SysID & SS.ref & SS.hyp & SS: ref$>$1 & SS: hyp$>$1 & SS: ref$>$1 \& hyp$>$1 \\ 
  \hline
fbk & 0.036 & 0.040 & 0.450 & 0.615 & 0.065 \\ 
  kit & 0.017 & 0.025 & 0.254 & 0.800 & 0.054 \\ 
  mitll & 0.019 & 0.026 & 0.321 & 0.714 & 0.036 \\ 
  naist & 0.019 & 0.026 & 0.336 & 0.715 & 0.051 \\ 
  nict & 0.018 & 0.025 & 0.305 & 0.763 & 0.069 \\ 
  prke & 0.039 & 0.042 & 0.488 & 0.585 & 0.073 \\ 
  rwth & 0.023 & 0.032 & 0.310 & 0.737 & 0.047 \\ 
  uedin & 0.025 & 0.032 & 0.317 & 0.700 & 0.017 \\ 
   \hline
\end{tabular}
}
\caption{Left: Percentage of reference/hypothesis words appearing in a substitution span. Right: Percentage of substitution spans containing multiple reference words, multiple hypothesis words, or both.} 
\label{tbl:span-stats}
\end{table}

We can corroborate this by
observing in Table~\ref{tbl:span-stats} that,
across all ASR systems,
$70.4\% (\pm 2.5\%)$ 
of the substitution spans involve multiple hypothesis words, while only
$34.8\% (\pm 2.8\%)$
contain multiple reference words.
The first figure may be explained by the presence of out-of-vocabulary words in the ASR reference, as well as the effects of domain variation in the evaluation data.
The alignment of multiple reference words to a single hypothesis word may be indicators of mispronunciations and/or underarticulation by the speaker.
These hypotheses should be explored in future work.

\subsection{Word classes and morphology}
\label{sec:asr-wc}

%

\begin{table}[tb]
\centering
{\footnotesize
\begin{tabular}{lrr}
  \hline
ErrorType & WER & POWER \\ 
  \hline
D.closed & 0.112 & 0.097 \\ 
  D.open & 0.067 & 0.041 \\ 
  I.closed & 0.101 & 0.065 \\ 
  I.open & 0.096 & 0.033 \\ 
  S.closed\_closed.morph & 0.001 & 0.001 \\ 
  S.closed\_closed.nomorph & 0.147 & 0.139 \\ 
  S.closed\_open.nomorph & 0.069 & 0.036 \\ 
  S.open\_closed.morph & 0.001 & 0.001 \\ 
  S.open\_closed.nomorph & 0.106 & 0.069 \\ 
  S.open\_open.morph & 0.052 & 0.054 \\ 
  S.open\_open.nomorph & 0.247 & 0.165 \\ 
  SS.closed\_span &  & 0.010 \\ 
  SS.open\_span &  & 0.186 \\ 
  SS.span\_closed &  & 0.019 \\ 
  SS.span\_open &  & 0.069 \\ 
  SS.span\_span &  & 0.016 \\ 
   \hline
\end{tabular}
}
\caption{Proportion of ASR error types by word class, averaged by ASR system. Substitution labels (S, SS) show the alignment from reference class to hypothesis class. Substitution spans (SS) contain a \textit{span} of words aligned either to a single word or another span. Word-level substitutions (S) are classified as morphological/non-morphological errors.} 
\label{tbl:propErrorsTable1}
\end{table}


Given the inconsistent error labeling in WER, which types of errors are actually being skewed by false alignments?
To answer this question, we annotate the reference and hypothesis words by their word class and observe their alignment statistics.
We additionally apply lemmatization to distinguish morphological errors from other substitution types.
According to the word statistics in Table~\ref{tbl:word-stats},
the ratio of open to closed class words remains the same across each ASR hypothesis and the reference (gold).
The proportion of errors associated with each ASR error type is shown in Table~\ref{tbl:propErrorsTable1}.

\subsubsection{Word-level substitution errors.}
Both WER and POWER report that the majority of substitution errors are within the same word class.
We observe that the number of closed-closed class substitutions remain the same between WER and POWER 
and that the proportion of open-open class substitutions caused by morphological errors remains the same.
POWER reports significantly fewer cross-class substitution errors, 
part of which are attributed to the correction of misalignments.
We also observe a 8\% reduction in open-open class substitution errors, 
which are most often replaced with substitution spans containing the word-level substitution error and one or more short function words
(e.g. \textit{Brown in} from Fig.~\ref{fig:power-misalign}).

\subsubsection{Deletions and Insertions.}
According to WER, deletion and insertion errors account for 
$37.7\% (\pm 0.7\%)$ of all errors.
WER marks nearly as many open class insertions as closed class insertions,
but suggests that closed class deletions are more prominent than open class ones
($6.7\% \pm 0.4\%$ open class deletions versus $11.2\% \pm 0.5\%$ closed class deletions).
However, with POWER, deletion and insertion errors only account for 
$23.6\% (\pm 0.8\%)$ of all errors,
with the majority of the reduction attributed to fewer open class insertion errors
($3.3\% \pm 0.1\%$).

\begin{table}[tb]
\footnotesize
\centering
\begin{tabular}{ll|ll}
\hline
\multicolumn{2}{c|}{WER} & \multicolumn{2}{c}{POWER}      \\
Reference    & Hypothesis & Reference  & Hypothesis       \\
\hline
a             & today   & a day        & today           \\
ascending     & and     & ascending    & and sending     \\
anesthetize   & and     & anesthetize  & and decent size \\
butchering    & the     & butchering   & maturing        \\
centigrade    & cents   & centigrade   & cents a great   \\
crude         & crudely & crude leaf   & crudely         \\
cyclones      & soy     & cyclones     & soy clones      \\
face-to-face  & face    & face-to-face & face to face    \\
of            & obama   & of anatomic  & obama panic    
\end{tabular}
\caption{Confusion pair examples using WER and POWER.}
\label{tbl:conf-pairs}
\end{table}

\subsubsection{Substitution spans.}
The majority of substitution spans have a single open class reference word
($18.6\% \pm 0.7\%$),
such as \textit{anatomy}\rightarrow\textit{and that to me} in Fig.~\ref{fig:power-misalign};
these represent the third most common POWER error type.
Likewise, the presence of a substitution span in the ASR reference indicates that the hypothesis word is likely to be a content word
($6.9\% \pm 0.8\%$).
Closed class function words are unlikely to be aligned to substitution spans
($2.9\% \pm 0.3\%$),
due to the fact that most have few syllables and thus cannot easily be mistaken with multiple words.
Instead, as indicated in Table \ref{tbl:propErrorsTable1}, closed class words are more likely to be missing from the corresponding output (i.e. deletion or insertion errors).

Table~\ref{tbl:conf-pairs} provides confusion pair examples that demonstrate the utility of substitution span alignments.
While WER often contains noisy confusion pairs, such as $\textit{a}\rightarrow\textit{today}$, 
POWER identifies that the error was actually a misrecognition of \textit{a day} as \textit{today}, since many American speakers pronounce today as /t ax d ey/, and thus the error is likely an insertion of the /t/ phoneme.

\bibliographystyle{IEEEbib}
\bibliography{paper}

\begin{thebibliography}{10}

\bibitem{Matusov:06}
E.~Matusov, S.~Kanthak, and H.~Ney,
\newblock ``Integrating speech recognition and machine translation: Where do we
  stand?,''
\newblock in {\em Proceedings of ICASSP}, Toulouse, France, 2006, pp.
  1217--1220.

\bibitem{Bertoldi:07}
N.~Bertoldi, R.~Zens, and M.~Federico,
\newblock ``Speech translation by confusion network decoding,''
\newblock in {\em Proceedings of ICASSP}, Honolulu, HA, 2007, pp. 1297--1300.

\bibitem{Casacuberta:08}
Francisco Casacuberta, Marcello Federico, Hermann Ney, and Enrique Vidal,
\newblock ``Recent efforts in spoken language processing,''
\newblock {\em IEEE Signal Processing Magazine}, vol. 25, no. 3, pp. 80--88,
  May 2008.

\bibitem{Ruiz:14:AMTA}
Nicholas Ruiz and Marcello Federico,
\newblock ``{Assessing the Impact of Speech Recognition Errors on Machine
  Translation Quality},''
\newblock in {\em Association for Machine Translation in the Americas (AMTA)},
  Vancouver, Canada, 2014, pp. 261--274.

\bibitem{Cettolo:2013:IWSLT}
Mauro Cettolo, Jan Niehues, Sebastian St\"{u}ker, Luisa Bentivogli, and
  Marcello Federico,
\newblock ``{Report on the 10th IWSLT Evaluation Campaign},''
\newblock in {\em Proc. of the International Workshop on Spoken Language
  Translation}, December 2013.

\bibitem{Cettolo:2014:IWSLT}
Mauro Cettolo, Jan Niehues, Sebastian St\"{u}ker, Luisa Bentivogli, and
  Marcello Federico,
\newblock ``{Report on the 11th IWSLT Evaluation Campaign},''
\newblock in {\em Proceedings of the International Workshop on Spoken Language
  Trnaslation (IWSLT)}, Lake Tahoe, USA, December 2014.

\bibitem{Festival}
Alan~W. Black and Paul~A. Taylor,
\newblock ``The {F}estival {S}peech {S}ynthesis {S}ystem: System
  documentation,''
\newblock Tech. {R}ep. HCRC/TR-83, {H}uman {C}ommunciation {R}esearch {C}entre,
  {U}niversity of {E}dinburgh, Scotland, UK, 1997,
\newblock Avaliable at http://www.cstr.ed.ac.uk/projects/festival.html.

\bibitem{Bird:2009:NLP}
Steven Bird, Ewan Klein, and Edward Loper,
\newblock {\em Natural Language Processing with Python},
\newblock O'Reilly Media, Inc., 1st edition, 2009.

\bibitem{Searle:1974:mixed}
S.~R. Searle,
\newblock ``Prediction, mixed models, and variance components,''
\newblock Tech. {R}ep. BU-468-M, Biometrics Unit, Cornell University, June
  1973.

\bibitem{Snover:06}
Matthew Snover, Bonnie Dorr, Rich Schwartz, Linnea Micciulla, and John Makhoul,
\newblock ``A study of translation edit rate with targeted human annotation,''
\newblock in {\em 5th Conference of the Association for Machine Translation in
  the Americas (AMTA)}, Boston, Massachusetts, August 2006.

\bibitem{R:2013}
{{R Core Team}},
\newblock {\em {R: A Language and Environment for Statistical Computing}},
\newblock {R Foundation for Statistical Computing}, {Vienna, Austria}, {2013}.

\bibitem{R:lme4}
Douglas Bates, Martin Maechler, Ben Bolker, and Steven Walker,
\newblock {\em lme4: Linear mixed-effects models using Eigen and S4}, 2014,
\newblock R package version 1.1-6.

\bibitem{Goldwater:10}
Sharon Goldwater, Daniel Jurafsky, and Christopher~D. Manning,
\newblock ``Which words are hard to recognize? prosodic, lexical, and
  disfluency factors that increase speech recognition error rates,''
\newblock {\em Speech Communication}, vol. 52, no. 3, pp. 181--200, 2010.

\bibitem{White:10}
Christopher White, Geoffrey Zweig, Lukas Burget, Petr Schwarz, and Hynek
  Hermansky,
\newblock ``Confidence estimation, oov detection and language id using
  phone-to-word transduction and phone-level alignments,''
\newblock in {\em Proceedings of ICASSP}, 2008.

\bibitem{DBLP:conf/icassp/TamLZW14}
Yik{-}Cheung Tam, Yun Lei, Jing Zheng, and Wen Wang,
\newblock ``{ASR} error detection using recurrent neural network language model
  and complementary {ASR},''
\newblock in {\em {IEEE} International Conference on Acoustics, Speech and
  Signal Processing, {ICASSP} 2014, Florence, Italy, May 4-9, 2014}. 2014, pp.
  2312--2316, {IEEE}.

\bibitem{Tsvetkov:14}
Yulia Tsvetkov, Florian Metze, and Chris Dyer,
\newblock ``Augmenting translation models with simulated acoustic confusions
  for improved spoken language translation.,''
\newblock in {\em EACL}, 2014, pp. 616--625.

\bibitem{Ruiz:15:Interspeech}
Nicholas Ruiz, Qin Gao, William Lewis, and Marcello Federico,
\newblock ``{Adapting Machine Translation Models toward Misrecognized Speech
  with Text-to-Speech Pronunciation Rules and Acoustic Confusability},''
\newblock in {\em Proceedings of Interspeech}, Dresden, Germany, September
  2015, ISCA.

\end{thebibliography}
\end{document}


%
\maketitle
\begin{abstract}
We propose a variation to the commonly used Word Error Rate (WER) metric for speech recognition evaluation which incorporates the alignment of phonemes, in the absence of time boundary information.
After computing the Levenshtein alignment on words in the reference and hypothesis transcripts, spans of adjacent errors are converted into phonemes with word and syllable boundaries and a phonetic Levenshtein alignment is performed.
The phoneme alignment information is used to correct the word alignment labels in each error region.
We demonstrate that our Pho\-net\-i\-cal\-ly-Oriented Word Error Rate (POWER) yields similar scores to WER with the added advantages of better word alignments and the ability to capture one-to-many alignments corresponding to homophonic errors in speech recognition hypotheses.
These improved alignments allow us to better trace the impact of Levenshtein error types in speech recognition on downstream tasks such as speech translation.
\end{abstract}
\begin{keywords}
automatic speech recognition, speech translation, mixed-effects models, error analysis
\end{keywords}
\section{Introduction}
\label{sec:intro}

Spoken language translation (SLT) systems are comprised by, at minimum, two components: 
an automatic speech recognition (ASR) system which provides audio transcripts of source language utterances,
and a machine translation (MT) system that translates the transcripts.
While there have been a number of efforts to construct tightly-coupled ASR and MT systems that are jointly trained and optimized,
the majority of SLT systems employ a cascading approach in which ASR systems are trained and evaluated independently from the MT system \cite{Matusov:06,Bertoldi:07,Casacuberta:08}.
In such a training paradigm, it is not clear how the introduction of ASR errors will affect translation quality.
While there is a high correlation between ASR errors and translation quality, 
the impact of various ASR error types is still an open research problem.

ASR performance is typically evaluated using the Word Error Rate (WER) metric, which labels errors as word-level substitutions, deletions, or insertions, based on the Levenshtein word alignment between a reference transcript and an ASR hypothesis.
Since the Levenshtein aligner often must decide between several ``optimal'' labeling sequences according to its objective function, 
it may select an alignment sequence that does not adhere to phonetic or linguistic relationships between the word types.
Although some evaluation toolkits use timestamp information to guide the alignment process, the problem persists.
For example, the mapping of \textit{a$\rightarrow$doctor} in Fig.~\ref{fig:wer-alignment} has the side-effects of aligning \textit{Dr.$\rightarrow$brahmin} and deleting \textit{Stanford}, thereby misaligning three content words.
Additionally, WER is not capable of identifying homophonic errors across word spans, such as \textit{anatomy$\rightarrow$and that to me}\footnote{We use the term ``homophonic'' to indicate groups of word sequences that are phonetically similar, but not necessarily identical, to one another.}.

These weaknesses in the word-level Levenshtein alignments used by WER inhibit the use of linguistically annotated ASR errors in assessing the quality of downstream speech-centric tasks, such as speech translation.
In response, we introduce an additional step in the alignment process, 
which computes pho\-net\-i\-cal\-ly-oriented word alignments across adjacent word errors that were predicted by WER.
We employ the text analysis component of a text-to-speech (TTS) engine, which dictactes written text based on a pronunciation dictionary, letter-to-sound rules, and context-dependent pronunciation rules for numbers, ordinals, and acronyms.

\begin{figure*}[t]
\footnotesize
\centering
\begin{tabular}{|c|c|}
\hline 
\textbf{WER} & \textbf{POWER} \\
\hline 
\begin{minipage}{0.40\textwidth}
\begingl
\gla traditional way  of  learning  human  {}   {}    {}  anatomy //
	\glb traditional way  of  loaning   human  and  that  to  me //
	\glc {}          {}   {}  S         {}     I    I     I   S //
	\endgl
\end{minipage} 
& 
	\begin{minipage}{0.40\textwidth}
\begingl
\gla traditional way  of  learning  human  anatomy //
	\glb traditional way  of  loaning   human  ``{and that to me}'' //
	\glc {}          {}   {}  S         {}     {SS (1:4)} //
	\endgl
\end{minipage} 
\\ 
\hline

\begin{minipage}{0.40\textwidth}
\begingl
\gla we  developed  with  a       Dr.      Brown  in  Stanford  //
	\glb we  developed  with  doctor  brahmin  stamp  or  {}        //
	\glc {}  {}         {}    S       S        S      S   D         //
	\endgl
\end{minipage} 
& 
	\begin{minipage}{0.40\textwidth}
\begingl
\gla we  developed  with  a   Dr.      ``{Brown  in}''  Stanford       //
	\glb we  developed  with  {}  doctor   brahmin          ``{stamp or}'' //
	\glc {}  {}         {}    D   S        {SS (2:1)}       {SS (1:2)}     //
	\endgl
\end{minipage} 
\\  
\hline 

%

%
\end{tabular} 
\caption{Error alignment differences between WER and POWER. POWER aligns homophonic errors such as \textit{anatomy $\rightarrow$ and~that~to~me}, while WER rate only aligns single words (e.g. \textit{anatomy$\rightarrow$me}).}
\label{fig:wer-alignment}
\end{figure*}

In this paper, we 
describe the application of pho\-net\-i\-cal\-ly-oriented word alignment on spans of adjacent errors 
and show that these corrected alignments significantly alter the distribution of ASR word error types.
We demonstrate its utility in text normalization as a pre-ASR evaluation step and
additionally apply the re-aligned error types to a SLT error analysis experiment that measures the impact of speech recognition errors on SLT quality.

\section{Pho\-net\-i\-cal\-ly-Oriented Word Alignment}
The ambiguity in the word-level Levenshtein aligner is centered around the placement of substitution errors in an alignment sequence.
As shown in Fig.~\ref{fig:wer-alignment}, the error spans contain at least one substitution error and a number of insertion or deletion errors.

Our pho\-net\-i\-cal\-ly-oriented word alignment algorithm is divided into two stages.
First, we capture error spans whose error labels are likely to be ambiguous.
The reference and hypothesis words in each span are transcribed into phonemes by a TTS analyzer.
Each phoneme is treated as an independent token and word and syllable boundary tokens are introduced.
The reference and hypothesis tokens are aligned using a variant of the Levenshtein alignment algorithm that introduces the following constraints:
\begin{enumerate}
\item Boundary tokens may not be substituted.
\item Vowel phonemes can only be aligned to other vowels (including r-colored vowels, but not semivowels).
\item Consonant phonemes can only be aligned to other consonants (including semivowels).
\end{enumerate}
The boundary tokens provide an implicit distance constraint, penalizing adjacent phonemes within the same syllable when they are aligned far from one another.

In the second stage, we recombine the phonetic alignments into word alignments by performing a left-to-right scan of the alignment sequence.
Substitution alignments are identified by considering the words covered by the aligned phonemes contained between two ``correct''-aligned word boundary markers in the reference and hypothesis.
Single word substitutions (S) are distinguished from 
substitution spans (SS) containing multiple words in the reference or the hypothesis.
If a sequence of reference phonemes are terminated with a word boundary, but no hypothesis words have been scanned, the reference word is marked as a deletion (D). 
Likewise, a hypothesis word with no aligned reference word is marked as an insertion (I).

Returning to Fig.~\ref{fig:wer-alignment}, the Levenshtein aligner used in WER could have alternatively aligned the reference word \textit{anatomy} to any one of the hypothesis words currently marked as insertion errors.
However, \textit{anatomy} is pronounced similarly to the entire sequence of the four hypothesis words in the error span.
The pho\-net\-i\-cal\-ly-oriented alignment in Fig.~\ref{fig:phoneme-alignment} captures this phenomenon by aligning the smallest word boundary closure across the entire span of reference and hypothesis words, thereby identifying \textit{anatomy$\rightarrow$and~that~to~me} as a substitution span and provides the alignment on the right-hand side of Fig.~\ref{fig:wer-alignment}.
Likewise, 
while WER may have considered slightly better word alignments like \textit{Brown$\rightarrow$brahmin} and \textit{Stanford$\rightarrow$or}, it is incapable of capturing relationships such as \textit{Stanford$\rightarrow$ stamp~or}.

\begin{figure}[ht]
\footnotesize
\centering
\begin{tabular}{|c|}
\hline
\begin{minipage}{0.38\textwidth}
\begingl
\gla $\|$     \#  ax  n  {} {}      \#  {}  ae  t  {}      \#  {} ax  m  {}      \#  {} iy  $\|$ //
\glb $\|$     \#  ae  n  d  $\|$    \#  dh  ae  t  $\|$    \#  t  ax  {} $\|$    \#  m  iy  $\|$ //
\glc {}       {}  S   {} I  I       {}  I   {}  {} I       {}  I  {}  D  I       {}  I  {}  {} //
\endgl 
\end{minipage}
\\
\hline
\end{tabular}
\caption{
\label{fig:phoneme-alignment}
Pho\-net\-i\-cal\-ly-oriented alignment of \textit{anatomy} to \textit{and~that~to~me}, with word ($\|$) and syllable (\#) boundaries.}
\end{figure}

\subsection{Word alignment heuristics}
While the pho\-net\-i\-cal\-ly-oriented alignments provide better pho\-net\-i\-cal\-ly-grounded alignments, its underlying Levenshtein alignment algorithm must also decide between multiple equally-weighted best paths.
\usetikzlibrary{matrix}

\begin{figure}[tb]
\footnotesize
\centering

\begin{tikzpicture}
  \matrix (m) [matrix of math nodes,row sep=0.75em,column sep=0.75em,minimum width=0.75em]
  {
   &    & \| & \# & ao & l  & \| & \# & ae & t  & \| \\
   & 0  & 3  & 6  & 9  & 12 & 15 & 18 & 21 & 24 & 27 \\
\| & 3  & 0  & 3  & 6  & 9  & 12 & 15 & 18 & 21 & 24 \\
\# & 6  & 3  & 0  & 3  & 6  & 9  & 12 & 15 & 18 & 21 \\
ao & 9  & 6  & 3  & 0  & 3  & 6  & 9  & 12 & 15 & 18 \\
r  & 12 & 9  & 6  & 3  & 4  & 7  & 10 & 13 & 16 & 19 \\
\| & 15 & 12 & 9  & 6  & 7  & 4  & 7  & 10 & 13 & 16 \\};
  \begin{scope}[every node/.style={scale=.75}]
  \path[-stealth]
    (m-2-3) edge [dashed] (m-2-2)
    (m-2-4) edge [dashed] (m-2-3)
    (m-2-5) edge [dashed] (m-2-4)
    (m-2-6) edge [dashed] (m-2-5)
    (m-2-7) edge [dashed] (m-2-6)
    (m-2-8) edge [dashed] (m-2-7)
    (m-2-9) edge [dashed] (m-2-8)
    (m-2-10) edge [dashed] (m-2-9)
    (m-2-11) edge [dashed] (m-2-10)
    
    (m-3-2) edge [dashed] (m-2-2)
    (m-3-3) edge [double,blue] node [above] {1} (m-2-2)
    (m-3-4) edge [dashed] (m-3-3)
    (m-3-5) edge [dashed] (m-3-4)
    (m-3-6) edge [dashed] (m-3-5)
    (m-3-7) edge [dashed] (m-3-6)
            edge [dashed] (m-2-6)
    (m-3-8) edge [dashed] (m-3-7)
    (m-3-9) edge [dashed] (m-3-8)
    (m-3-10) edge [dashed] (m-3-9)
    (m-3-11) edge [dashed] (m-3-10)
            edge [dashed] (m-2-10)
    
    (m-4-2) edge [dashed] (m-3-2)
    (m-4-3) edge [dashed] (m-3-3)
    (m-4-4) edge [double,blue] node [above] {1} (m-3-3)
    (m-4-5) edge [dashed] (m-4-4)
    (m-4-6) edge [dashed] (m-4-5)
    (m-4-7) edge [dashed] (m-4-6)
    (m-4-8) edge [dashed] (m-4-7)
            edge [dashed] (m-3-7)
    (m-4-9) edge [dashed] (m-4-8)
    (m-4-10) edge [dashed] (m-4-9)
    (m-4-11) edge [dashed] (m-4-10)
    
    (m-5-2) edge [dashed] (m-4-2)
    (m-5-3) edge [dashed] (m-4-3)
    (m-5-4) edge [dashed] (m-4-4)
    (m-5-5) edge [double,blue] node [above] {1} (m-4-4)
    (m-5-6) edge [red] node [above] {1} (m-5-5)
    (m-5-7) edge [red] node [above] {1} (m-5-6)
    (m-5-8) edge [red] node [above] {1} (m-5-7)
    (m-5-9) edge [red] node [above] {1} (m-5-8)
    (m-5-10) edge [dashed] (m-5-9)
    (m-5-11) edge [dashed] (m-5-10)
    
    (m-6-2) edge [dashed] (m-5-2)
    (m-6-3) edge [dashed] (m-5-3)
    (m-6-4) edge [dashed] (m-5-4)
    (m-6-5) edge [dashed] (m-5-5)
    (m-6-6) edge [double,blue] node [above] {1} (m-5-5)
    (m-6-7) edge [red] node [above] {1} (m-6-6)
    (m-6-8) edge [red] node [above] {1} (m-6-7)
    (m-6-9) edge [red] node [above] {1} (m-6-8)
    (m-6-10) edge [red] node [above] {1} (m-6-9)
             edge [red] node [above] {1} (m-5-9)
    (m-6-11) edge [dashed] (m-6-10)
    
    (m-7-2) edge [dashed] (m-6-2)
    (m-7-3) edge [dashed] (m-6-3)
    (m-7-4) edge [dashed] (m-6-4)
    (m-7-5) edge [dashed] (m-6-5)
    (m-7-6) edge [dashed] (m-6-6)
    (m-7-7) edge [double,blue] node [above] {1} (m-6-6)
    (m-7-8) edge [double,blue] node [above] {0} (m-7-7)
    (m-7-9) edge [double,blue] node [above] {0} (m-7-8)
    (m-7-10) edge [double,blue] node [above] {0} (m-7-9)
    (m-7-11) edge [red] node [above] {1} (m-6-10)
             edge [double,blue] node [above] {0} (m-7-10)
    ;
    \end{scope}

\end{tikzpicture}

\resizebox{\columnwidth}{!}{\begin{tabular}{|c|c|c|}
\hline 
\begin{minipage}{0.17\textwidth}
\begingl
\gla 
$\|$  \#  ao   l   $\|$  \#  ae   t   $\|$ //
\glb 
$\|$  \#  ao   {}  {}       {}  {}   r   $\|$ //
\glc 
{}       {}  {}   D   D        D   D    S   {}      //
	\endgl 
{\centering \textcolor{red}{incorrect} \\}
\end{minipage} 
&
\begin{minipage}{0.17\textwidth}
\begingl
\gla 
$\|$  \#  ao   l   $\|$  \#  ae  t   $\|$ //
\glb 
$\|$  \#  ao   r   {}       {}  {}  {}  $\|$ //
\glc 
{}       {}  {}   S   D        D   D   D   {}      //
	\endgl 
{\centering \textcolor{red}{incorrect} \\}
\end{minipage} 
& 
\begin{minipage}{0.17\textwidth}
\begingl
\gla 
$\|$  \#  ao   l   $\|$  \#  ae   t  $\|$ //
\glb
$\|$  \#  ao   r   $\|$  {}  {}   {} {}       //
\glc 
{}       {}  {}   S   {}       D   D    D  D      //
	\endgl 
{\centering \textcolor{blue}{correct} \\}
\end{minipage} 
\\ 
\hline
\end{tabular}}
\caption{POWER alignments for \textit{all at}$\rightarrow$\textit{or}. The Levenshtein backtrack matrix shows three alignments with the same edit distance scores. 
The third and correct alignment (highlighted in the backtrack matrix) compactly aligns \textit{all}$\rightarrow$\textit{or}, while the others greedily align \textit{or} to multiple reference words.}
\label{fig:power-misalign}
\end{figure}

In particular, for alignments with large differences in the number of reference and hypothesis syllables, 
our implementation tends to align the first and last word boundaries close to the beginning and end of the alignment sequence. 
For example, Fig.~\ref{fig:power-misalign} shows three candidate alignments for the error span \textit{all at}$\rightarrow$\textit{or} that minimize the edit distance. Two out of three alignments attempt to align \textit{or} to the entire two-syllable reference.
However, only \textit{all} should align to \textit{or} as a substitution, and \textit{at} should be considered a deletion error.
We resolve ambiguities like these by finding the alignment that minimizes the number of alignment gaps between the first and last word boundaries in both the reference and hypothesis.
In practice, we do this by encoding the best paths in the Levenshtein backtrack matrix into an edge-weighted graph and use Dijkstra's algorithm to find the best path.

Since there still remains some noise in the phonetic alignments, we introduce a couple of heuristics to prevent the aligner from overzealously marking single-syllable words as members of a substitution span, when in reality they do not have a phonetic correspondence on the other side.
When annotating a substitution span, we keep a record of the number of reference and hypothesis syllables.
If 
there is an extra syllable in the reference or hypothesis, we check if it is the first syllable of a new word. 
If so, we mark this word as a deletion or insertion error, respectively.

\subsection{Scoring}
Our Pho\-net\-i\-cal\-ly-Oriented Word Error Rate (POWER) score is defined nearly identically to WER as:
\begin{align}
\textrm{POWER} = \frac{S+D+I+SS}{L}, \nonumber \\
SS = \sum_{span} \textrm{max}(\left\vert{span_{ref}}\right\vert, \left\vert{span_{hyp}}\right\vert),
\label{eq:power}
\end{align}
where $L$ is the length of the reference and $S$, $D$, and $I$ are the number of word-level substitution, deletion, and insertion labels, respectively.
$SS$ is the count of substitution spans, weighted by the maximum number of words in each span.
These one-to-many or many-to-many word alignments indicate phonetic confusability as the cause of the error.


\section{Experiments}
Following the experimental framework of \cite{Ruiz:14:AMTA},  
we perform our experiments on an intersection of the ASR and MT results of the IWSLT 2013 evaluation campaign \cite{Cettolo:2013:IWSLT}, which focused on the translation of TED talks.
These 580 utterances map a subset of the ASR hypotheses 
provided by 8 ASR systems
to the corresponding MT inputs in the English-French MT track.
The unpunctuated MT input serves as the ASR reference data.
Eight French human post-edited references serve as the MT references in the SLT analysis.

In order to minimize the effects of formatting issues on our experimental results, the ASR hypotheses are evaluated, normalized, recased, and punctuated according to the MT input and are translated by a baseline English-French Moses SMT system, corresponding to the \textrm{WIT}$^3$ data from 2014 \cite{Cettolo:2014:IWSLT}.
Since we desire to keep the ASR reference intact, we apply oracle-based text normalization and punctuation insertion techniques similar to that of \cite{Ruiz:14:AMTA}, instead of applying a general .glm normalization file to the both the ASR reference and hypothesis.
We use POWER to align and normalize hypothesis words with respect to the reference.
POWER uses the Festival TTS system with the CMU English pronunciation dictionary \cite{Festival} to convert words into phonemes. 
Prior to other normalization steps, we use the \textsc{varcon} tool from SCOWL\footnote{https://github.com/kevina/wordlist} to convert British English words in the ASR hypotheses to American English.
We also use libraries from NLTK \cite{Bird:2009:NLP} to annotate ASR errors with part-of-speech and word class information, as well as lemmatization for morphological analysis.

\begin{table}[tb]
\centering
\footnotesize
\resizebox{\columnwidth}{!}{\begin{tabular}{l|rrrr|rr|rHrH}
\hline
          & \multicolumn{4}{c}{Data set}                  & \multicolumn{2}{c}{ASR WER $\downarrow$}  & \multicolumn{4}{c}{SLT} \\
System    & tokens         & open        & closed & ratio & orig & norm & BLEU $\uparrow$ & METEOR $\uparrow$ & TER $\downarrow$  & LENGTH \\
\hline
fbk       & 10095          & 5581        & 4514   & 1.24  & 21.3     & 16.5 & 51.9 & 62.0   & 38.5 & 99.7   \\
kit       & 10141          & 5571        & 4570   & 1.22  & 15.1     & 10.1 & 55.4 & 65.4   & 35.2 & 99.9   \\
mitll     & 10147          & 5594        & 4553   & 1.23  & 16.3     & 11.4 & 55.0 & 64.7   & 35.8 & 99.9   \\
naist     & 10076          & 5571        & 4505   & 1.24  & 15.6     & 10.5 & 55.1 & 65.3   & 35.3 & 99.4   \\
nict      & 10165          & 5595        & 4570   & 1.22  & 14.4     & 9.2  & 56.5 & 66.4   & 34.3 & 99.8   \\
prke      & 10106          & 5545        & 4561   & 1.22  & 21.2     & 16.5 & 52.1 & 62.2   & 38.4 & 99.6   \\
rwth      & 10160          & 5563        & 4597   & 1.21  & 16.4     & 11.6 & 54.3 & 64.4   & 36.2 & 100.0  \\
uedin     & 10151          & 5592        & 4559   & 1.23  & 17.1     & 12.3 & 54.6 & 64.5   & 36.1 & 99.9   \\
\hline
gold      & 10158          & 5614        & 4544   & 1.24  & 0.0      & 0.0  & 62.9 & 71.8   & 29.1 & 100.1 
\end{tabular}}
\caption{Statistics for each ASR system on
the ratio of open to closed class words by ASR system; 
ASR WER scores before and after text normalization; 
and English-French translation scores for normalized and punctuated ASR hypotheses,
compared to the ASR reference (gold).}
\label{tbl:word-stats}
\end{table}

We conduct two sets of experiments. 
First, we analyze the ASR error annotations given by the WER and POWER alignments to measure the effects of Levenshtein alignment heuristics on the reported results.
In the second set of experiments, we construct several mixed-effects models \cite{Searle:1974:mixed} that measure the contribution of various ASR error types on MT errors. 
Table~\ref{tbl:word-stats} provides summary statistics on the words in each ASR hypothesis, the WER scores before and after text normalization for each ASR output, and scores from two translation metrics.
In particular, the POWER-driven text normalization reduces each system's WER scores by 5\%.

\section{ASR Error Analysis}
\label{sec:asr-eval}
In this set of experiments, we observe the contribution of particular error types to the global WER and POWER scores for each ASR system.
We outline the shortcomings of WER's statistics due to the erratic behavior of Levenshtein aligners.


\begin{knitrout}
\definecolor{shadecolor}{rgb}{0.969, 0.969, 0.969}\color{fgcolor}\begin{figure}

{\centering \includegraphics[width=\maxwidth]{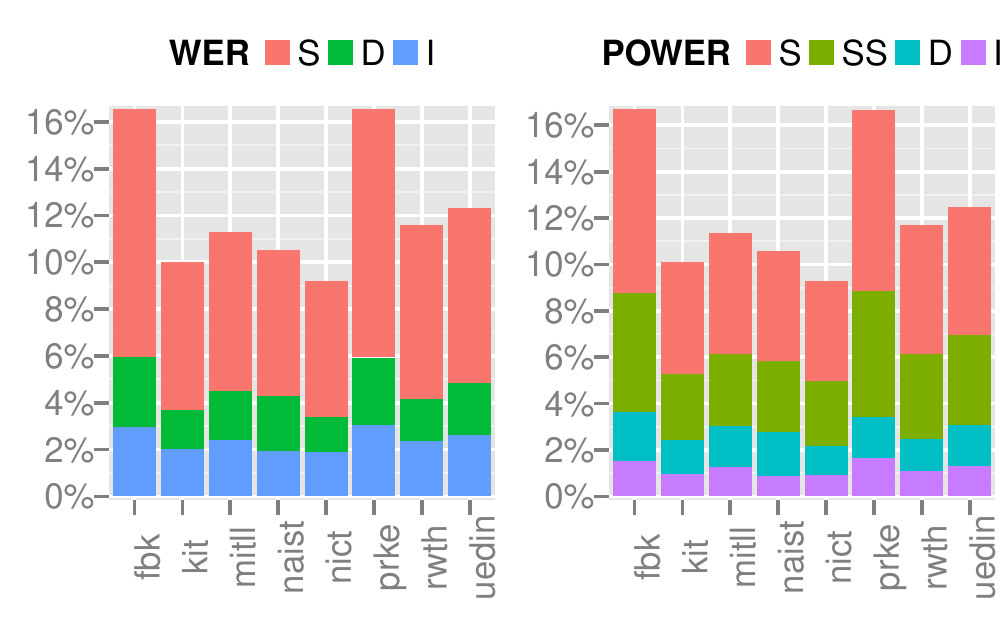} 

}

\caption[Distribution of error types for WER (left) and POWER (right) for each IWSLT 2013 ASR evaluation participant]{Distribution of error types for WER (left) and POWER (right) for each IWSLT 2013 ASR evaluation participant.}\label{fig:asrErrorSummaryPlot}
\end{figure}

\end{knitrout}

\subsection{Basic Levenshtein error types}
\label{sec:asr-eval:basic}

We begin by looking at the basic ASR error types (S, D, I, and SS), which implicitly contain no linguistic information.
Fig.~\ref{fig:asrErrorSummaryPlot} shows the contribution of the basic Levenshtein error types toward the error rate score for each ASR system.
%
%
According to WER, substitutions intuitively make up the majority of error types 
($62.3\% \pm 0.7\%$). 
Across all ASR systems, WER suggests that the number of deletions are slightly lower than the number of insertion errors
($17.9\% \pm 0.7\%$ deletions
and $19.8\% \pm 0.5\%$ insertions).

However, POWER suggests that
roughly half of 
these alleged insertion errors ($10.0\% \pm 0.3\%$)
are instances where a larger reference word is being hypothesized as a homophonic sequence of shorter words.
Likewise, a portion of ``deletion'' errors are instances where multiple reference words were hypothesized as a longer homophonic word
($4.1\% \pm 0.5\%$).
%
%
Since these substitution span errors are typically cases of one-to-many alignments, the number of reported word-level substitution errors are reduced.
As such, POWER claims that
$30.0\%$ $(\pm 0.7\%)$ of the errors are substitution spans involving homophony, 
leaving $13.8\% (\pm 0.8\%)$ of the remaining errors as deletions
and only $9.8\% (\pm 0.3\%)$ as insertions whose pronunciations do not align to any words -- both measures are substantially lower than those reported by WER.
The remaining
$46.4\% (\pm 0.5\%)$ 
are word-level substitutions.

\begin{table}[tb]
\centering
{\footnotesize
\begin{tabular}{l|rr|rrr}
  \hline
SysID & SS.ref & SS.hyp & SS: ref$>$1 & SS: hyp$>$1 & SS: ref$>$1 \& hyp$>$1 \\ 
  \hline
fbk & 0.036 & 0.040 & 0.450 & 0.615 & 0.065 \\ 
  kit & 0.017 & 0.025 & 0.254 & 0.800 & 0.054 \\ 
  mitll & 0.019 & 0.026 & 0.321 & 0.714 & 0.036 \\ 
  naist & 0.019 & 0.026 & 0.336 & 0.715 & 0.051 \\ 
  nict & 0.018 & 0.025 & 0.305 & 0.763 & 0.069 \\ 
  prke & 0.039 & 0.042 & 0.488 & 0.585 & 0.073 \\ 
  rwth & 0.023 & 0.032 & 0.310 & 0.737 & 0.047 \\ 
  uedin & 0.025 & 0.032 & 0.317 & 0.700 & 0.017 \\ 
   \hline
\end{tabular}
}
\caption{Left: Percentage of reference/hypothesis words appearing in a substitution span. Right: Percentage of substitution spans containing multiple reference words, multiple hypothesis words, or both.} 
\label{tbl:span-stats}
\end{table}

We can corroborate this by
observing in Table~\ref{tbl:span-stats} that,
across all ASR systems,
$70.4\% (\pm 2.5\%)$ 
of the substitution spans involve multiple hypothesis words, while only
$34.8\% (\pm 2.8\%)$
contain multiple reference words.
The first figure may be explained by the presence of out-of-vocabulary words in the ASR reference, as well as the effects of domain variation in the evaluation data.
The alignment of multiple reference words to a single hypothesis word may be indicative of mispronunciations and/or underarticulation by the speaker.
These hypotheses should be explored in future work.

\subsection{Word classes and morphology}
\label{sec:asr-wc}

%

\begin{table}[tb]
\centering
{\footnotesize
\begin{tabular}{lrrrr}
  \hline
ErrorType & WER & POWER & WER Rank & POWER Rank \\ 
  \hline
S.open\_open & 0.299 & 0.219 & 1 & 1 \\ 
  SS.open\_span &  & 0.186 &  & 2 \\ 
  S.closed\_closed & 0.148 & 0.140 & 2 & 3 \\ 
  D.closed & 0.112 & 0.097 & 4 & 4 \\ 
  S.open\_closed & 0.107 & 0.069 & 4 & 6 \\ 
  SS.span\_open &  & 0.069 &  & 6 \\ 
  I.closed & 0.101 & 0.065 & 6 & 7 \\ 
  D.open & 0.067 & 0.041 & 8 & 8 \\ 
  S.closed\_open & 0.069 & 0.036 & 8 & 8 \\ 
  I.open & 0.096 & 0.033 & 6 & 10 \\ 
  SS.span\_closed &  & 0.019 &  & 12 \\ 
  SS.span\_span &  & 0.016 &  & 12 \\ 
  SS.closed\_span &  & 0.010 &  & 13 \\ 
   \hline
\end{tabular}
}
\caption{Proportion of ASR error types by word class, averaged across all ASR systems and ranked by importance. Substitution labels (S, SS) show the alignment from reference class to hypothesis class. Substitution spans (SS) contain a \textit{span} of words aligned either to a single word or another span.} 
\label{tbl:propErrorsTable1}
\end{table}


Given the inconsistent error labeling in WER, which types of errors are actually being skewed by false alignments?
To answer this question, we annotate the reference and hypothesis words by their word class and observe their alignment statistics.
We additionally apply lemmatization to distinguish morphological errors from other substitution types.
According to the word statistics in Table~\ref{tbl:word-stats},
the ratio of open to closed class words remains the same across each ASR hypothesis and the reference (gold).
The proportion of errors associated with each ASR error type is shown in Table~\ref{tbl:propErrorsTable1}.

\textbf{Word-level substitution errors.}
Both WER and POWER report that the majority of substitution errors are within the same word class.
While the proportion of closed-closed class substitutions remain the same,
POWER reports 
8\% 
fewer open-open class substitution errors, 
which are often instances of substitution error spans containing a word-level substitution error and one or more short function words
(e.g.~\textit{Brown in}$\rightarrow$ \textit{brahmin} from Fig.~\ref{fig:power-misalign}).
Of the open-open class substitution errors,
5.4\%
are morphological errors.
POWER likewise reports 
7\%
fewer cross-class substitution errors, 
many of which are attributed to the correction of misalignments.

\textbf{Deletions and Insertions.}
According to WER, deletion and insertion errors account for 
$37.7\% (\pm 0.7\%)$ of all errors.
WER marks nearly as many open class insertions as closed class insertions,
but suggests that closed class deletions are more prominent than open class ones
($6.7\% \pm 0.4\%$ open versus $11.2\% \pm 0.5\%$ closed class deletions).
However, with POWER, deletion and insertion errors only account for 
$23.6\%$ $(\pm 0.8\%)$ of all errors,
with the majority of the reduction attributed to fewer open class insertion errors
($3.3\% \pm 0.1\%$).
An example of a corrected open class ``deletion'' is \textit{Stanford}$\rightarrow$ \textit{stamp or} from Fig.~\ref{fig:power-misalign}.

\textbf{Substitution spans.}
The majority of substitution spans have a single open class reference word
($18.6\% \pm 0.7\%$),
such as \textit{anatomy}$\rightarrow$\textit{and that to me} in Fig.~\ref{fig:power-misalign};
these represent the second most common POWER error type.
Likewise, the presence of a substitution span in the ASR reference indicates that the hypothesis word is likely to be a content word
($6.9\% \pm 0.8\%$).
Closed class function words are unlikely to be aligned to substitution spans
($2.9\% \pm 0.3\%$),
since most have few syllables that cannot easily be mistaken for multiple words.
Instead, as shown in Table \ref{tbl:propErrorsTable1}, closed class words are more likely to be deletion or insertion errors. 

\begin{table}[tb]
\footnotesize
\centering
\begin{tabular}{ll|ll}
\hline
\multicolumn{2}{c|}{WER} & \multicolumn{2}{c}{POWER}      \\
Reference    & Hypothesis & Reference  & Hypothesis       \\
\hline
a             & today   & a day        & today           \\
ascending     & and     & ascending    & and sending     \\
anesthetize   & and     & anesthetize  & and decent size \\
butchering    & the     & butchering   & maturing        \\
centigrade    & cents   & centigrade   & cents a great   \\
crude         & crudely & crude leaf   & crudely         \\
cyclones      & soy     & cyclones     & soy clones      \\
face-to-face  & face    & face-to-face & face to face    \\
of            & obama   & of anatomic  & obama panic     \\
\hline
\end{tabular}
\caption{Confusion pair examples using WER and POWER.}
\label{tbl:conf-pairs}
\end{table}

Table~\ref{tbl:conf-pairs} provides confusion pair examples from FBK's ASR system output that demonstrate the utility of POWER.
Word confusion pairs such as \textit{a}$\rightarrow$\textit{today} are likely errors induced by an ASR language model that biases the acoustic model to artificially recognize non-existant phonemes.
Likewise, POWER is able to provide insight that \textit{crude}$\rightarrow$\textit{crudely} is not a morphological error, but rather another language model-induced bias that considers \textit{leaf} an unlikely successor to \textit{crude}.
Other confusion pairs include word normalizations, affix errors, and phonetic confusions.

\section{SLT Error Analysis}
\label{sec:slt-eval}

Given that POWER yields a significantly different distribution of error types, how can it be leveraged to understand the impact of ASR errors on downstream natural language processing tasks? 
We turn our attention to the translation of TED talks from English to French.
Similar to \cite{Ruiz:14:AMTA}, we measure the impact 
of utterance-level ASR errors on their associated translation score
by measuring the increase in translation error rate ($\Delta\textrm{TER}$) \cite{Snover:06} against a gold standard translation which contains no ASR errors.
We use linear mixed-effects regression models 
to measure the importance of each ASR error type,
taking into consideration random effects caused by an ASR system and the particular features of each ASR utterance.
We use the R \cite{R:2013} implementation of mixed-effects models in the \textit{lme4} library \cite{R:lme4}.
All of our models are fit using maximum likelihood and incorporate random intercepts for each ASR utterance (labeled as \textit{UttID}) and ASR system (labeled as \textit{SysID}), as well as a random slope by the WER score.
We use the repeated observations of our 580 speech utterances by eight ASR systems, yielding a total of 4,640 observations.
As fixed effects, we normalize the counts of each ASR error type by the length of the ASR reference for each utterance in order to consider its contribution toward the utterance-level WER score.
In each model, \textit{SysID} was not significant, with a standard deviation near zero.


\begin{table}[tb] \centering 
  \caption{Fixed effects coefficients and 95\% confidence intervals for the first three mixed-effects models, which measure the effect of ASR error types on $\Delta\textrm{TER}$ for English-French SLT. The baseline encapsulates all error types in a single WER measure, while the subsequent models use WER and POWER-aligned error types.} 
  \label{tbl:lmerSideBySide} 
\footnotesize 
\begin{tabular}{@{\extracolsep{0pt}}lD{.}{.}{-3} D{.}{.}{-3} D{.}{.}{-3} } 
\\[-1.8ex]\hline 
\hline \\[-1.8ex] 
 & \multicolumn{1}{c}{WER} & \multicolumn{1}{c}{WER.basic} & \multicolumn{1}{c}{POWER.basic} \\ 
\\[-1.8ex] & \multicolumn{1}{c}{(1)} & \multicolumn{1}{c}{(2)} & \multicolumn{1}{c}{(3)}\\ 
\hline \\[-1.8ex] 
 WER & 0.630^{***} &  &  \\ 
  & \multicolumn{1}{c}{(0.586$, $0.674)} &  &  \\ 
  WER.D &  & 0.564^{***} & 0.615^{***} \\ 
  &  & \multicolumn{1}{c}{(0.506$, $0.622)} & \multicolumn{1}{c}{(0.556$, $0.674)} \\ 
  WER.I &  & 0.707^{***} & 0.829^{***} \\ 
  &  & \multicolumn{1}{c}{(0.642$, $0.772)} & \multicolumn{1}{c}{(0.753$, $0.906)} \\ 
  WER.S &  & 0.624^{***} & 0.649^{***} \\ 
  &  & \multicolumn{1}{c}{(0.578$, $0.671)} & \multicolumn{1}{c}{(0.601$, $0.696)} \\ 
  WER.SS &  &  & 0.535^{***} \\ 
  &  &  & \multicolumn{1}{c}{(0.487$, $0.584)} \\ 
  Constant & 0.001 & 0.001 & -0.0001 \\ 
  & \multicolumn{1}{c}{(-0.003$, $0.004)} & \multicolumn{1}{c}{(-0.002$, $0.004)} & \multicolumn{1}{c}{(-0.003$, $0.003)} \\ 
 \hline \\[-1.8ex] 
Observations & \multicolumn{1}{c}{4,640} & \multicolumn{1}{c}{4,640} & \multicolumn{1}{c}{4,640} \\ 
Log Likelihood & \multicolumn{1}{c}{6,172.170} & \multicolumn{1}{c}{6,180.631} & \multicolumn{1}{c}{6,194.288} \\ 
Akaike Inf. Crit. & \multicolumn{1}{c}{-12,330.340} & \multicolumn{1}{c}{-12,343.260} & \multicolumn{1}{c}{-12,368.580} \\ 
Bayesian Inf. Crit. & \multicolumn{1}{c}{-12,285.240} & \multicolumn{1}{c}{-12,285.280} & \multicolumn{1}{c}{-12,304.150} \\ 
\hline 
\hline \\[-1.8ex] 
\textit{Note:}  & \multicolumn{3}{r}{$^{*}$p$<$0.1; $^{**}$p$<$0.05; $^{***}$p$<$0.01} \\ 
\end{tabular} 
\end{table}

\subsection{WER versus POWER features}
Our first experiments consider the ASR error labels provided by WER and POWER.
Our baseline considers a single error feature, corresponding to the WER score for each utterance.
We compare it to two mixed-effects models and report their coefficients in Table \ref{tbl:lmerSideBySide}.
\textit{WER.basic} is trained with WER's basic substitution (WER.S), deletion (WER.D), and insertion (WER.I) labels;
\textit{POWER.basic} is trained additionally with POWER's substitution span labels (WER.SS).
Both sets of features are normalized by the reference length in order to be a decomposition of the WER metric. 

As in \cite{Ruiz:14:AMTA}, we observe a significant difference between \textit{WER.basic} and the baseline, rejecting the null hypothesis that each basic ASR error type contibutes equally to translation quality, in terms of $\Delta\textrm{TER}$ 
($\chi^2(2)=16.922, p < 2.12 \times 10^{-4}$).
We additionally observe a significant difference between the standard WER-aligned error types (\textit{WER.basic}) and the POWER-aligned error types (\textit{POWER.basic}) that include substitution spans
($\chi^2(1)=27.314, p=1.73 \times 10^{-7}$),
indicating that substitution spans are a significant predictor of translation quality.
As shown in Table \ref{tbl:lmerSideBySide}, while the impact of substitution errors remains in principle the same, the impact of insertions increase sharply, both due to the higher quality of the error labels and their lower frequency.
\textit{POWER.basic} indicates that an utterance with a WER (or equivalently, POWER) score of $10\%$ as insertion errors would expect an increase in TER by $0.1 \times 0.829 - 0.0001 = 8.3\%$, while $10\%$ in substitution errors would correspond to a TER increase of $0.1 \times 0.649 - 0.0001 = 6.49\%$.

\subsection{Frequency-weighted ASR errors}
While the coefficients in Table \ref{tbl:lmerSideBySide} show the expected increase in TER for each percentage of WER associated with a particular error type, 
an error type with a high coefficient but a low frequency may not be important from an error correction standpoint.
Ideally, we wish to measure which ASR errors are particularly problematic for a given SLT task.

We construct an additional mixed-effects model using the word class-annotated error types from Section \ref{sec:asr-wc}.
We compute frequency-weighted scores for each error type, based on the observations in our data set.
Considering the fixed and random effect scores on each utterance, we measure the average weighted contribution of each ASR error type toward the $\Delta\textrm{TER}$ measure.
In other words, if we observe one ASR error of a particular type, how much is it expected to degrade the translation quality?
By doing so, we seek to rank the importance of each error type.
Table \ref{tbl:interpretCoefficients_wordclass} reports the mean and standard error for each weighted error type, using the word class-annotated error types provided by POWER.
We observe that, similar to the ASR-only experiments, within-class substitution errors have the highest frequency-weighted contribution toward $\Delta\textrm{TER}$.
While substitution spans containing open class reference words have a high weighted score, substitution spans with open class hypothesis words have a substantially lower weighted score.

\begin{table}[tb]
\centering
{\footnotesize
\begin{tabular}{lrrrr}
  \hline
ErrorType & coef & weight-mean & weight-se & SLT Rank \\ 
  \hline
WER.S.open\_open & 0.687 & 0.0175 & 0.0006 & 1 \\ 
  WER.S.closed\_closed & 0.585 & 0.0132 & 0.0005 & 2 \\ 
  WER.SS.open\_span & 0.723 & 0.0123 & 0.0006 & 3 \\ 
  WER.D.closed & 0.451 & 0.0069 & 0.0004 & 4 \\ 
  WER.I.closed & 0.548 & 0.0059 & 0.0004 & 6 \\ 
  WER.S.open\_closed & 0.663 & 0.0057 & 0.0003 & 6 \\ 
  WER.D.open & 0.546 & 0.0048 & 0.0005 & 7 \\ 
  WER.I.open & 0.553 & 0.0044 & 0.0004 & 9 \\ 
  WER.S.closed\_open & 0.590 & 0.0036 & 0.0002 & 9 \\ 
  WER.SS.span\_open & 0.802 & 0.0038 & 0.0003 & 9 \\ 
  WER.SS.span\_closed & 0.757 & 0.0016 & 0.0002 & 12 \\ 
  WER.SS.span\_span & 1.036 & 0.0015 & 0.0003 & 12 \\ 
  WER.SS.closed\_span & 0.713 & 0.0011 & 0.0002 & 13 \\ 
  (Intercept) & 0.000 & 0.0002 & 0.0003 & 14 \\ 
   \hline
\end{tabular}
}
\caption{Mixed-effects coefficients (coef) for POWER ASR error types with word class annotations, and their mean frequency-weighted contributions toward translation $\Delta\textrm{TER}$ (weight-mean).} 
\label{tbl:interpretCoefficients_wordclass}
\end{table}

\section{Discussion}
Based on the error statistics provided above and recorded in Table~\ref{tbl:propErrorsTable1}, we identify the following error types as interesting to focus on when constructing models to cope with ASR errors.
$16.2\% (\pm 0.6\%)$ of the ASR errors are either insertions or deletions on closed class words.
These types are also ranked highly in our SLT experiments.
While a recovery model to insert or delete hypothesis words is non-trivial, we consider closed class words to be low-hanging fruit, since the number of alternative words are small and a language model would likely have statistics that support their inclusion or removal.
While closed class words can be under-articulated, they receive the majority of their support from the language model due to the large amount of observations.
The advantage of using POWER to model errors is that we have more confidence that the words we mark as deletions or insertions during training/development are really unaligned words. 

Likewise, it is useful to consider the effects of substitution spans, as they are among the most frequent errors caused by ASR systems.
By identifying substitution error spans, we are able to capture a consecutive string of words that can significantly alter the meaning of a sentence.
However, oftentimes they are due to homophonic errors
where an ASR system may have reasonable confidence in the phonemes detected, but due to the interaction between the acoustic and language models, a shorter sequence of similar-sounding words was selected.
It would be worthwhile to identify common phonetic error patterns to either rescore ASR hypotheses or carry forward the ambiguity of a span of words in the hypothesis to allow the downstream process to decide which similar-sounding alternative makes the most sense.

\section{Related Work}
Mixed-effects models were first used in ASR error analysis in \cite{Goldwater:10} to analyze the effects of lexical, prosodic, contextual, and disfluency features of individual words on WER. They show that of the various disfluency types, word fragments, non-final repetitions, and words preceding fragments have a significant impact on WER.
Our work proposes a pho\-net\-i\-cal\-ly-oriented word alignment process that is more successful in aligning words of the same word class. Such an alignment process would alter the individual WER measure proposed in \cite{Goldwater:10}, which could provide more reliable results.

A related area of work is ASR confidence estimation, which seeks to label erroneous words in an ASR hypothesis.
\cite{White:10} uses the comparison of phones in a strong ASR system and a weak ASR system without a language model as features for error detection. Regions where the difference is large indicate a higher likelihood of errors.
Other approaches include using ASR consnsus votes as well as recurrent neural networks to capture longer contexts \cite{DBLP:conf/icassp/TamLZW14}.


On downstream tasks such as speech translation, \cite{Tsvetkov:14,Ruiz:15:Interspeech} propose ASR channel modeling techniques that rely on the concept of phonetic confusability to convert error-free source language phrases into ASR-like outputs in order to model ASR errors during machine translation model training. 
Pho\-net\-i\-cal\-ly-oriented alignments could be used in either approach to identify error regions during training to focus the channel model on confusable words.
Our use of a TTS analyzer to generate pronunciation sequences on ASR references and hypotheses is based on \cite{Ruiz:15:Interspeech}.

\section{Conclusion}
We have developed a pho\-net\-i\-cal\-ly-oriented word alignment pipeline as an extension to Word Error Rate's Levenshtein aligner.
Spans of adjacent Levenshtein errors containing minimally one substitution error are converted into phonemes with word and syllable boundaries. A second Levenshtein alignment process on phonemes is carried out and the alignment information is used to guide the word alignment process.
We demonstrate that our pho\-net\-i\-cal\-ly-oriented word alignments generate virtually the same error rate score as WER, with the added benefit of more reliable substitution error tags, and a reduction of erroneous deletion and insertion error labels on open class words.
We demonstrate that the use of pho\-net\-i\-cal\-ly-oriented error labels significantly alters the statistics gathered from error analyses on ASR outputs.
Additionally, for speech translation error analysis tasks, we demonstrate that our pho\-net\-i\-cal\-ly-oriented word error alignments result in better error models in mixed-effects modeling.
We demonstrate that homophonic error spans comprise a significant portion of ASR errors with a large impact on speech translation quality and deserve to be considered as an additional substitution error type in error recovery efforts.
Our POWER software is available as open source software for the research community at 
\mbox{https://github.com/NickRuiz/power-asr}.

\bibliographystyle{IEEEbib}
\bibliography{paper}


\section{SLT Error Analysis}
\label{sec:slt-eval}

\begin{table}[tb] \centering 
  \caption{Fixed effects coefficients and 95\% confidence intervals for the first three mixed effects models, which measure the effect of ASR error types on $\Delta\textrm{TER}$ for English-French SLT. The baseline encapsulates all error types in a single WER measure, while the subsequent models use WER and POWER-aligned error types.} 
  \label{tbl:lmerSideBySide} 
\footnotesize 
\begin{tabular}{@{\extracolsep{0pt}}lD{.}{.}{-3} D{.}{.}{-3} D{.}{.}{-3} } 
\\[-1.8ex]\hline 
\hline \\[-1.8ex] 
 & \multicolumn{1}{c}{WER} & \multicolumn{1}{c}{WER.basic} & \multicolumn{1}{c}{POWER.basic} \\ 
\\[-1.8ex] & \multicolumn{1}{c}{(1)} & \multicolumn{1}{c}{(2)} & \multicolumn{1}{c}{(3)}\\ 
\hline \\[-1.8ex] 
 WER & 0.630^{***} &  &  \\ 
  & \multicolumn{1}{c}{(0.586$, $0.674)} &  &  \\ 
  WER.D &  & 0.564^{***} & 0.615^{***} \\ 
  &  & \multicolumn{1}{c}{(0.506$, $0.622)} & \multicolumn{1}{c}{(0.556$, $0.674)} \\ 
  WER.I &  & 0.707^{***} & 0.829^{***} \\ 
  &  & \multicolumn{1}{c}{(0.642$, $0.772)} & \multicolumn{1}{c}{(0.753$, $0.906)} \\ 
  WER.S &  & 0.624^{***} & 0.649^{***} \\ 
  &  & \multicolumn{1}{c}{(0.578$, $0.671)} & \multicolumn{1}{c}{(0.601$, $0.696)} \\ 
  WER.SS &  &  & 0.535^{***} \\ 
  &  &  & \multicolumn{1}{c}{(0.487$, $0.584)} \\ 
  Constant & 0.001 & 0.001 & -0.0001 \\ 
  & \multicolumn{1}{c}{(-0.003$, $0.004)} & \multicolumn{1}{c}{(-0.002$, $0.004)} & \multicolumn{1}{c}{(-0.003$, $0.003)} \\ 
 \hline \\[-1.8ex] 
Observations & \multicolumn{1}{c}{4,640} & \multicolumn{1}{c}{4,640} & \multicolumn{1}{c}{4,640} \\ 
Log Likelihood & \multicolumn{1}{c}{6,172.170} & \multicolumn{1}{c}{6,180.631} & \multicolumn{1}{c}{6,194.288} \\ 
Akaike Inf. Crit. & \multicolumn{1}{c}{-12,330.340} & \multicolumn{1}{c}{-12,343.260} & \multicolumn{1}{c}{-12,368.580} \\ 
Bayesian Inf. Crit. & \multicolumn{1}{c}{-12,285.240} & \multicolumn{1}{c}{-12,285.280} & \multicolumn{1}{c}{-12,304.150} \\ 
\hline 
\hline \\[-1.8ex] 
\textit{Note:}  & \multicolumn{3}{r}{$^{*}$p$<$0.1; $^{**}$p$<$0.05; $^{***}$p$<$0.01} \\ 
\end{tabular} 
\end{table}

Given that we have measured that POWER yields a significantly different distribution of error types, how can it be leveraged to understand the impact of ASR errors on downstream tasks, such as spoken language translation (SLT)?
We next turn our attention to the translation of TED talks from English to French.
Similar to \cite{Ruiz:14}, we measure the impact of particular ASR error types on SLT metrics, such as TER by measuring the increase in translation error rate $\Delta\textrm{TER}$ against a gold standard translation which contains no ASR errors.
We use linear mixed-effects models, which have been effectively used on linguistic data by \cite{Baayen:2008}, to measure the importance of each ASR error type. 
Mixed-effects models allow us to take into consideration random effects caused by an ASR system and the particular features of each ASR utterance.
We use the R \cite{R:2013} implementation of linear mixed-effects models in the \textit{lme4} library \cite{R:lme4}.
For all of our models, we provide random intercepts for the utterance (labeled as \textit{UttID}) and ASR system (labeled as \textit{SysID}), as well as a random slope by the WER score.
Unlike our ASR error analysis experiments, which considered the global WER score, the SLT task is concerned with the utterance-level WER and the associated translation error score.
The models are fit by maximum likelihood.
We use the repeated observations of our 580 speech utterances by eight ASR systems, yielding a total of 4,640 observations.
As fixed effects, we normalize the counts of each ASR error type by the length of the ASR reference for each utterance in order to consider its contribution toward the utterance-level WER score.
In each model, \textit{SysID} was not significant, with a standard deviation near zero.

\subsection{WER versus POWER features}
Our first experiments consider the ASR error labels provided through WER and POWER.
Our baseline considers a single error feature, corresponding to the WER score for each utterance.
We compare it to two mixed effects models, one trained with WER's basic substitution (WER.S), deletion (WER.D), and insertion (WER.I), labels, and the other trained additionally with POWER's substitution span labels (WER.SS).
Both sets of features are normalized by the reference length in order to be a decomposition of the WER metric.
The coefficients of the mixed effects models are listed in Table \ref{tbl:lmerSideBySide}.
As a baseline, we compare the model that rolls all ASR error types into the WER score to \textit{WER.basic}, which segments WER into substitution (WER.S), deletion (WER.D), and insertion (WER.I) labels.
As in \cite{Ruiz:14}, we observe a significant difference between \textit{WER.basic} and the baseline, rejecting the null hypothesis that each basic ASR error type contibutes equally to translation quality, in terms of $\DeltaTER$ 
($\chi^2(2)=16.922, p < \num{2.12 \times 10^{-4}}$).

Comparing our WER and POWER-trained models (\textit{WER.basic} and \textit{POWER.basic}, respectively), 
we observe a significant difference between the standard WER-aligned error types and the POWER-aligned error types that include substitution spans (WER.SS)
($\chi^2(1)=27.314, p=1.73 \times 10^{-7}$),
indicating that substitution spans are a significant predictor of translation quality.
As shown in Table \ref{tbl:lmerSideBySide}, while the impact of substitution errors remains in principle the same, the impact of insertions increased sharply, both due to the higher quality of the error labels and their lower frequency.
POWER.basic indicates that an utterance with a WER (or equivalently, POWER) score of $10\%$ as insertion errors would expect an increase in TER by $0.1 \times 0.829 - 0.0001 = 8.3\%$, while $10\%$ in substitution (WER.S) errors would correspond to a TER incrasee of $0.1 \times 0.649 - 0.0001 = 6.49\%$.

\begin{table}
\centering
\footnotesize

\begin{tabular}{lcc}
\hline
  & mean & \multicolumn{1}{c}{se} \\ 
\hline
WER.D.closed  & $0.00691$ & $0.00037$ \\
WER.D.open  & $0.00477$ & $0.00048$ \\
WER.I.closed  & $0.00591$ & $0.00039$ \\
WER.I.open  & $0.00441$ & $0.00038$ \\
WER.S.closed\_closed  & $0.01319$ & $0.00052$ \\
WER.S.closed\_open  & $0.00357$ & $0.00024$ \\
WER.S.open\_closed  & $0.00572$ & $0.00033$ \\
WER.S.open\_open  & $0.01746$ & $0.00061$ \\
WER.SS.closed\_span  & $0.00108$ & $0.00021$ \\
WER.SS.open\_span  & $0.01227$ & $0.00064$ \\
WER.SS.span\_closed  & $0.00156$ & $0.00024$ \\
WER.SS.span\_open  & $0.00377$ & $0.00032$ \\
WER.SS.span\_span  & $0.00153$ & $0.00027$ \\
\hline 
\end{tabular}

\caption{ASR error types and their mean frequency-weighted contribution toward translation $\Delta\textrm{TER}$.}
\label{tbl:interpretCoefficients_wordclass}
\end{table}

\subsection{Frequency-weighted ASR errors}
While the coefficients in Table \ref{tbl:lmerSideBySide} show the expected increase in TER for each percentage of WER associated with a particular error type, it does not provide an indication of how common an error will occur.
An error type with a high coefficient but a low frequency may not be important from an error correction standpoint.
Ideally, we wish to measure which ASR errors are particularly problematic for a given SLT task.

Focusing on word class-annotated error types, similar to Section \ref{sec:asr-wc}, we compute the frequency-weighted scores for each error type, based on the observations in our data set, and accounting for the random intercept by and random slope by WER and utterance.
Factoring the fixed and random effects for each utterance, we measure the average weighted contribution of each ASR error type toward the $\Delta\textrm{TER}$ measure.
By doing so, we seek to rank the importance of each error type.
Table \ref{tbl:interpretCoefficients_wordclass} reports the mean and standard error for each weighted error type, using the word class-annotated error types provided by POWER.
We observe that within-class substitution errors have the highest frequency-weighted contribution toward $\Delta\textrm{TER}$.

\bibliographystyle{IEEEbib}
\bibliography{paper}